\documentclass{article}
\usepackage{arxiv}

\usepackage[utf8]{inputenc} % allow utf-8 input
\usepackage[T1]{fontenc}    % use 8-bit T1 fonts
\usepackage{hyperref}       % hyperlinks
\usepackage{url}            % simple URL typesetting
\usepackage{booktabs}       % professional-quality tables
\usepackage{amsfonts}       % blackboard math symbols
\usepackage{nicefrac}       % compact symbols for 1/2, etc.
\usepackage{microtype}      % microtypography
\usepackage{lipsum}
\usepackage{float}
\usepackage{graphicx}
\usepackage{multirow}
\usepackage{array}
\usepackage{comment}
\usepackage{longtable}
\usepackage{ragged2e}
\usepackage{subfig}

\title{Machine learning based disease diagnosis: A comprehensive review}

\author{
  Md Manjurul Ahsan \\
  Industrial and Systems Engineering\\
  University of Oklahoma\\
  Norman, Oklahoma-73071 \\
  \texttt{ahsan@ou.edu} \\
   \And
 Zahed Siddique \\
  Department of Aerospace and Mechanical Engineering\\
  University of Oklahoma\\
  Norman, Oklahoma-73071\\
  \texttt{zsiddique@ou.edu}} 

\begin{document}
\maketitle

\begin{abstract}
Globally, there is a substantial unmet need to diagnose various diseases effectively. The complexity of the different disease mechanisms and underlying symptoms of the patient population presents massive challenges to developing the early diagnosis tool and effective treatment. Machine Learning (ML), an area of Artificial Intelligence (AI), enables researchers, physicians, and patients to solve some of these issues. Based on relevant research, this review explains how Machine Learning (ML) and Deep Learning (DL) are being used to help in the early identification of numerous diseases. To begin, a bibliometric study of the publication is given using data from the Scopus and Web of Science (WOS) databases. The bibliometric study of 1216 publications was undertaken to determine the most prolific authors, nations, organizations, and most cited articles. The review then summarizes the most recent trends and approaches in Machine Learning-based Disease Diagnosis (MLBDD), considering the following factors: algorithm, disease types, data type, application, and evaluation metrics. Finally, the paper highlights key results and provides insight into future trends and opportunities in the MLBDD area.
\end{abstract}

% keywords can be removed

\section*{Abbreviation}
ANN \quad Artificial Neural Network\\
CNN \quad Convolutional Neural Network\\
DL \quad Deep Learning\\
DNN \quad Deep Neural Networks\\
DT \quad Decision Trees\\
KMC \quad K-Means Clustering\\
KNN \quad K-Nearest Neighbors\\
LR  \quad Logistic Regression\\
ML  \quad Machine Learning\\
NN  \quad Neural Network\\
RF \quad Random Forest\\
SVM \quad Support Vector Machine\\
\section{Introduction}
Diagnosis is a branch of Artificial Intelligence (AI) focused with developing the algorithms and techniques capable of determining whether a system's behavior is correct. Medical diagnosis identifies the disease or conditions that explain a person's symptoms and signs. Typically, diagnostic information is gathered from the patient's history and physical examination~\cite{mcphee2010current}. It is frequently difficult due to the fact that many indications and symptoms are ambiguous and can only be diagnosed by trained health experts. Therefore, countries that lack enough health professionals for their populations, such as developing countries like Bangladesh and India, face difficulty providing proper diagnostic procedures for their maximum patients~\cite{ahsan2021detecting}.
Moreover, diagnosis procedures often require medical tests, which low-income people often find expensive and difficult to afford. \\
As humans are prone to error, it is not surprising that a patient may be over-diagnosed more often. If a person is over-diagnosed, a problem like unnecessary treatments will arise, harming individuals' health and economic waste~\cite{coon2014overdiagnosis}. According to the National Academics of Science, Engineering, and Medicine report of 2015, the majority of people will encounter at least one diagnostic mistake during their lifespan~\cite{balogh2015improving}.Various factors may influence the misdiagnosis, which includes:
\begin{itemize}
    \item lack of proper symptoms, which often unnoticeable
    \item the condition of rare disease
    \item the disease is omitted mistakenly from the consideration
\end{itemize}
Machine Learning (ML) is used practically everywhere, from cutting-edge technology (such as mobile phones, computers, and robotics) to health care (i.e., disease diagnosis, safety). ML is gaining traction in various fields, including disease diagnosis in health care. Many researchers and practitioners illustrate the promise of Machine Learning-based Disease Diagnosis (MLBDD), which is inexpensive and time-efficient~\cite{ahsan2021machine}. Traditional diagnosis processes are costly, time-consuming, and often require human intervention. While the individual's ability restricts traditional diagnosis techniques, ML-based systems have no such limitations, and machines do not get exhausted as humans do. As a result, a method to diagnose disease with outnumbered patients' unexpected presence in health care may be developed. To create MLBDD systems, health care data like images (i.e., X-ray, MRI) and tabular data (i.e., patients' conditions, age, and gender) are employed~\cite{ahsan2020deep}.\\
Machine Learning (ML) is a subset of AI that uses data as an input resource~\cite{stafford2020systematic}. The use of predetermined mathematical functions yields a result (classification or regression) that is frequently difficult for humans to accomplish. For example, using ML, locating malignant cells in a microscopic image is frequently simpler, which is typically challenging to conduct just by looking at the images. Furthermore, since advances in Deep Learning (a form of machine learning), the most current study shows MLBDD accuracy of above 90\%~\cite{ahsan2021machine}. Alzheimer's disease, Heart failure, Breast cancer, and Pneumonia are just a few of the diseases that may be identified with ML. The emergence of Machine Learning (ML) algorithms in disease diagnosis domains illustrates the technology's utility in medical fields. \\
Recent breakthroughs in ML difficulties, such as unbalanced data, ML interpretation, and ML ethics in medical domains, are only a few of the many challenging fields to handle in a nutshell~\cite{ahsan2020covid}. This paper gives a review that highlights the novel uses of ML and DL in disease diagnosis and provides an overview of development in this field in order to shed some light on this current trend, approaches, and issues connected with ML in disease diagnosis. It begins by outlining several methods to Machine Learning and Deep Learning techniques, and particular architecture for detecting and categorizing various forms of disease diagnosis.
\subsection{Motivation}
The purpose of this review is to provide insights to recent and future researchers and practitioners regarding Machine Learning-based Disease Diagnosis (MLBDD) that will aid and enable them to choose the most appropriate and superior Machine Learning/Deep Learning methods, thereby increasing the likelihood of rapid and reliable disease detection and classification in diagnosis. Additionally, the review aims at identifying potential studies related to the MLBDD. In general, the scope of this study is to provide the proper explanation for the following questions:
\begin{enumerate}
    \item What are some of the diseases that researchers and practitioners are particularly interested in when evaluating data-driven machine learning approaches?
    \item Which databases are incorporated into MLBDD?
    \item Which Machine Learning and Deep Learning approaches are presently used in health care to classify various forms of disease?
    \item Which architecture of Convolutional Neural Networks (CNNs) is widely employed in disease diagnosis?
    \item How is the model's performance evaluated? Is that sufficient?
\end{enumerate}
This paper summarizes the different Machine Learning (ML) and Deep Learning (DL) methods utilized in various disease diagnosis applications. The remainder of the paper is structured as follows: Section~\ref{background} discusses the background and overview of ML and DL, whereas Section~\ref{article} details the article selection technique. Section~\ref{bibliometric} includes bibliometric analysis. Section~\ref{ML} discusses the use of Machine Learning in various disease diagnoses, and Section~\ref{AD} identifies the most frequently utilized ML methods and datatypes based on the linked research. Section~\ref{discussion} discusses the findings, anticipated trends, and problems. Finally, Section~\ref{conclusion} concludes the article with a general conclusion.
\section{Basics and Background}\label{background}
\begin{figure}[htbp]
    \centering
    \includegraphics[width=\textwidth]{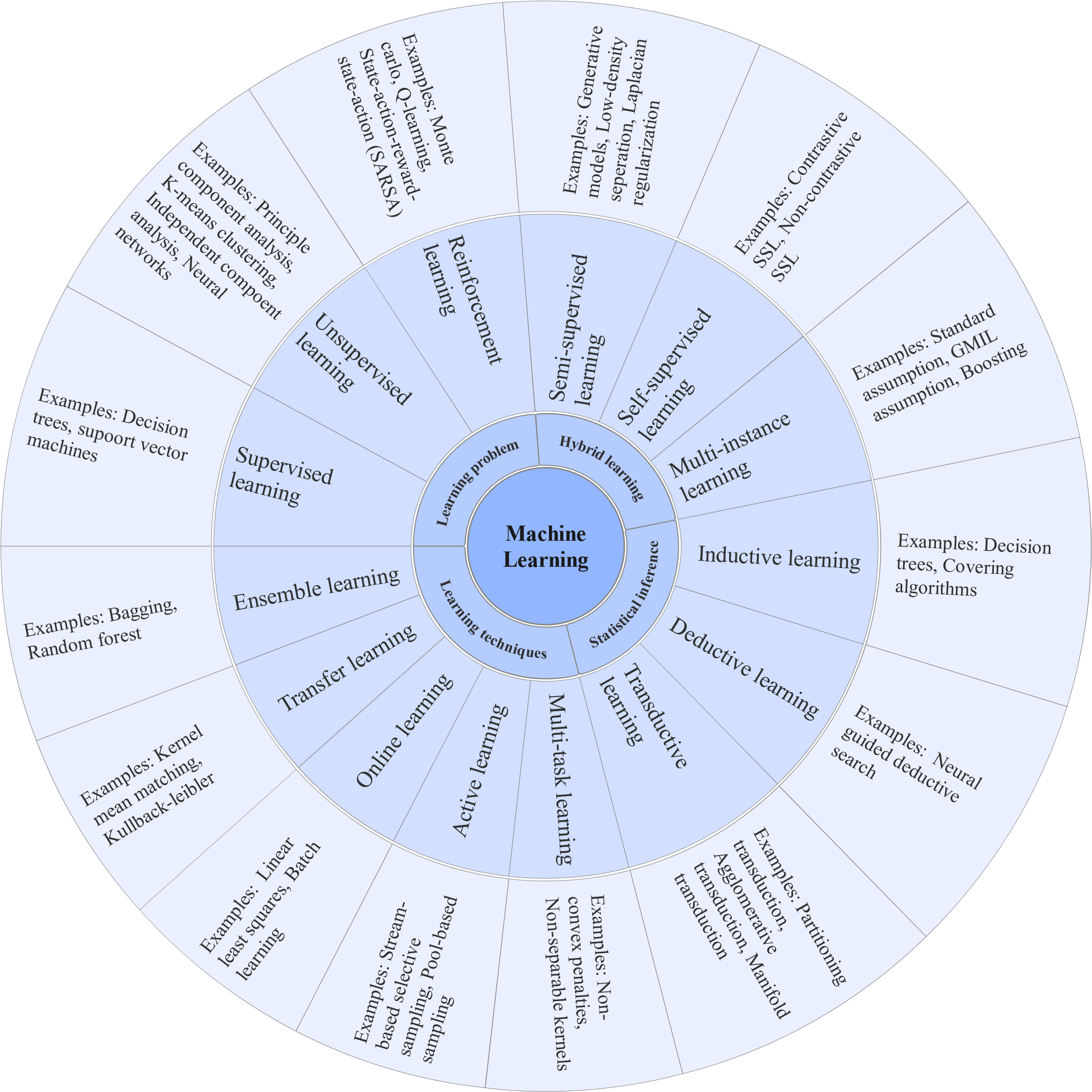}
    \caption{Different types of Machine Learning algorithms.}
    \label{fig:fig1}
\end{figure}

Machine Learning (ML) is an approach that analyzes data samples to create main conclusions using mathematical and statistical approaches, allowing machines to learn without programming. Arthur Samuel presented machine learning in games and pattern recognition algorithms to learn from experience in 1959, which was the first time the important advancement was recognized. The core principle of ML is to learn from data in order to forecast or make decisions depending on the assigned task~\cite{samuel1959some}. Thanks to Machine Learning (ML) technology, many time-consuming jobs may now be completed swiftly and with minimal effort. With the exponential expansion of computer power and data capacity, it is becoming simpler to train data-driven machine learning models to predict outcomes with near-perfect accuracy. Several papers offer various sorts of ML approaches~\cite{brownlee2016machine,houssein2021deep}.\\
The ML algorithms are generally classified into three categories such as supervised, unsupervised, and semi-supervised~\cite{brownlee2016machine}. However, ML algorithms can be divided into several subgroups based on different learning approaches, as shown in Fig.~\ref{fig:fig1}. Some of the popular ML algorithms include Linear Regression, Logistic Regression, Support Vector Machines (SVM), Random Forest (RF), and Naïve Bayes (NB)~\cite{brownlee2016machine}.
\subsection{Machine Learning Algorithms}
This section provides a comprehensive review of the most frequently used Machine Learning algorithms in disease diagnosis.
\subsubsection{Decision Tree}
The Decision Tree (DT) algorithm follows divide and conquer rules. In DT models, the attribute may take on various values known as classification trees; leaves indicate distinct classes, while branches reflect the combination of characteristics that result in those class labels. On the other hand, DT can take continuous variables called regression trees. C4.5 and EC4.5 are the two famous and most widely used DT algorithms~\cite{brijain2014survey}. DT is used extensively by following reference literature:~\cite{walse2020effective,rajendran2020predicting,tsao2018predicting,nurrohman2020parkinson}.
\subsubsection{Support Vector Machine}
For classification and regression-related challenges, Support Vector Machine (SVM) is a popular ML approach. SVM was introduced by Vapnik in the late 20$^{th}$ century~\cite{drucker1999support}. Apart from disease diagnosis, SVMs have been extensively employed in various other disciplines, including facial expression recognition, protein fold, distant homology discovery, speech recognition, and text classification. For unlabeled data, supervised ML algorithms are unable to perform. Using a hyperplane to find the clustering among the data, SVM can categorize unlabeled data. However, SVM output is not non-linearly separable. To overcome such problems, selecting appropriate kernel and parameters is two key factors when applying SVM in data analysis~\cite{houssein2021deep}.
\subsubsection{K-Nearest Neighbor}
K-Nearest Neighbor (KNN) classification is a non-parametric classification technique invented in 1951 by Evelyn Fix and Joseph Hodges. KNN is suitable for classification as well as regression analysis. The outcome of KNN classification is class membership. 
Voting mechanisms are used to classify the item. Euclidean distance techniques are utilized to determine the distance between two data samples. The projected value in regression analysis is the average of the values of the KNN~\cite{fix1989discriminatory}.
\subsubsection{Naïve Bayes}
The Naive Bayes (NB) classifier is a Bayesian-based probabilistic classifier. Based on a given record or data point, it forecasts membership probability for each class. The most probable class is the one having the greatest probability. Instead of predictions, the NB classifier is used to project likelihood~\cite{houssein2021deep}.
\subsubsection{Logistic Regression}
Logistic regression (LR) is an ML approach that is used to solve classification issues. The LR model has a probabilistic framework, with projected values ranging from 0 to 1. Examples of LR-based ML include spam email identification, online fraud transaction detection, and malignant tumor detection. The cost function, often known as the Sigmoid function, is used by LR. The Sigmoid function transforms every real number between 0 and 1~\cite{wright1995logistic}.
\subsubsection{AdaBoost}
Yoav Freund and Robert Schapire developed Adaptive Boosting, popularly known as AdaBoost. Adaboost is a classifier that combines multiple weak classifiers into a single classifier. AdaBoost works by giving greater weight to samples that are harder to classify and less weight to those that are already well categorized. It may be used for categorization as well as regression analysis~\cite{schapire2013explaining}.
\subsection{Deep Learning Overview}
Deep Learning (DL) is a sub-field of Machine Learning (ML) that employs multiple layers to extract both higher and lower-level information from input (i.e., images, numerical value, categorical values). The majority of contemporary DL models are built on Artificial Neural Networks (ANN), notably Convolutional Neural Networks (CNN), which may be integrated with other DL models, including generative models, Deep Belief Networks, and the Boltzmann Machine. 
Deep Learning may be classified into three types: supervised, semi-supervised, and unsupervised.. Deep Neural Networks (DNN), Reinforcement Learning, and Recurrent Neural Networks (RNN) are some of the most prominent DL architectures (RNN)~\cite{goodfellow2016deep}.\\
Each level in DL learns to convert its input data to the succeeding layers while learning distinct data attributes. For example, the raw input may be a pixel matrix in image recognition applications, and the first layers may detect the image's edges. On the other hand, the second layer will construct and encode the nose and eyes, and the third layer may recognize the face by merging all of the information gathered from the previous two layers~\cite{ahsan2020deep}.\\
In medical fields, DL has enormous promise. Radiology and pathology are two well-known medical fields that have widely used DL in disease diagnosis over the years~\cite{hayashi2019right}. Furthermore, collecting valuable information from molecular state and determining disease progression or therapy sensitivity are practical uses of DL that are frequently unidentified by human investigations~\cite{akkus2017deep}.\\
\subsubsection{Convolutional Neural Network}
\begin{figure}[htbp]
    \centering
    \includegraphics[width=\textwidth]{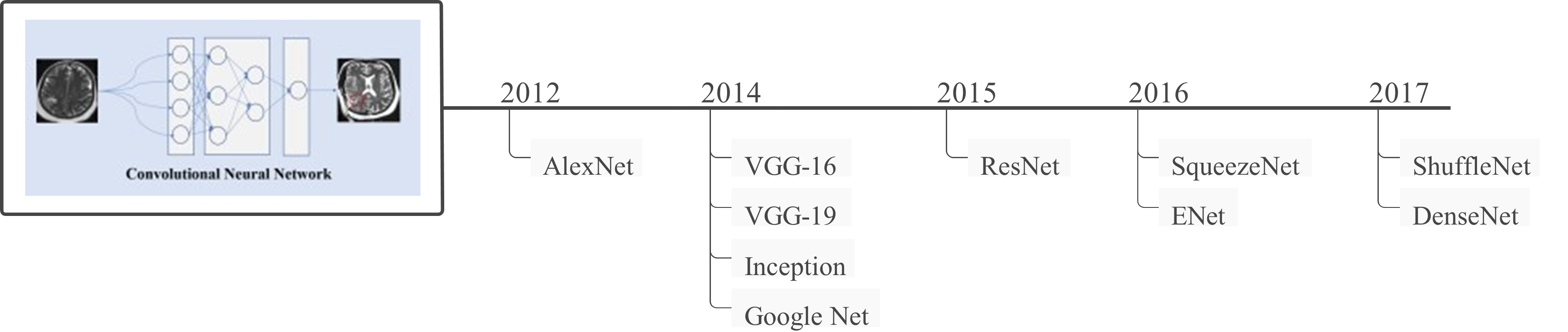}
    \caption{Some of the most well-known CNN models, along with their development time frames.}
    \label{fig:fig2}
\end{figure}

\begin{figure}[htbp]
    \centering
    \includegraphics[width=\textwidth]{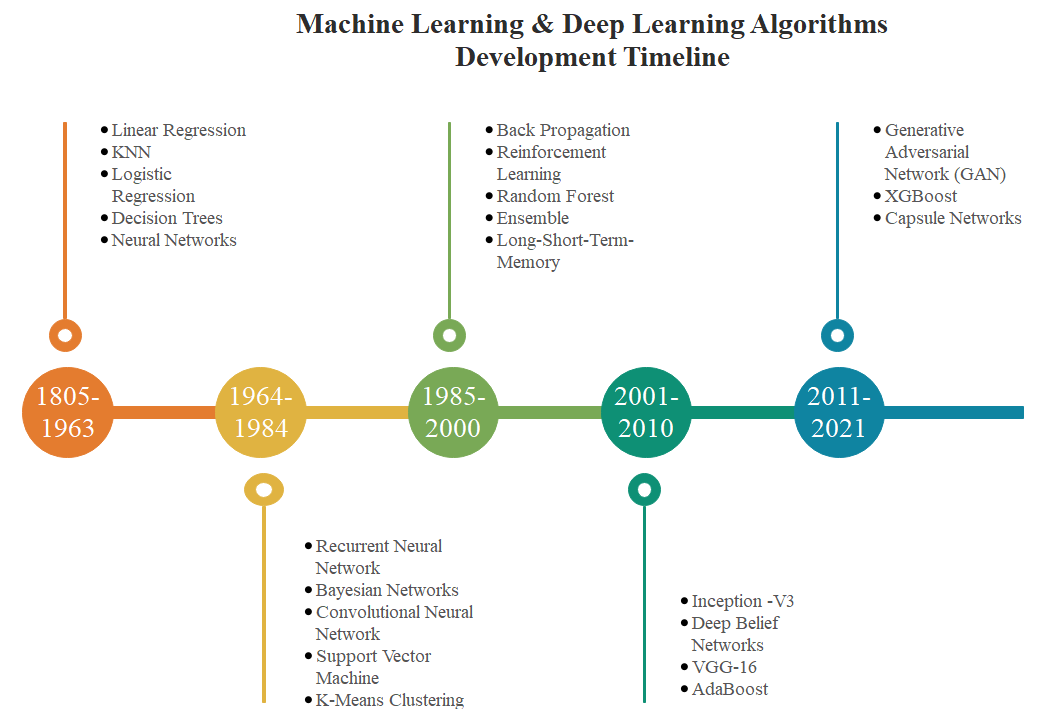}
    \caption{Illustration of Machine Learning and Deep Learning algorithms development timeline.}
    \label{fig:fig3}
\end{figure}

Convolutional Neural Networks (CNNs) are a subclass of Artificial Neural Networks (ANNs) that are extensively used in image processing.
CNN is widely employed in face identification, text analysis, human organ localization, and biological image detection or recognition~\cite{yap2017automated}. Since the initial development of CNN in 1989, a different type of CNN has been proposed that has performed exceptionally well in disease diagnosis over the last three decades. A CNN architecture comprises three parts: input layer, hidden layer, and output layer. The intermediate levels of any feed-forward network are known as hidden layers, and the number of hidden layers varies depending on the type of architecture. 
Convolutions are performed in hidden layers, which contain dot products of the convolution kernel with the input matrix. Each convolutional layer provides feature maps used as input by the subsequent layers. Following the concealed layer are more layers, such as pooling and fully connected layers~\cite{goodfellow2016deep}.
Several CNN models have been proposed throughout the years, and the most extensively used and popular CNN models are shown in Fig.~\ref{fig:fig2}.
\\
In general, it may be considered that ML and DL have grown substantially throughout the years. 
The increased computational capability of computers and the enormous number of data available inspire academics and practitioners to employ ML/DL more efficiently. A schematic overview of machine learning and deep learning algorithms and their development chronology is shown in Fig.~\ref{fig:fig3}, which may be a helpful resource for future researchers and practitioner
\subsection{Performance Evaluations}
This section describes the performance measures used in reference literature. Performance indicators, including accuracy, precision, recall, and f-1 score, are widely employed in disease diagnosis. For example, lung cancer can be categorized as true positive ($T_P$) or true-negative ($T_N$) if individuals are diagnosed correctly, while it can be categorized into false-positive ($F_P$) or false-negative ($F_N$) if misdiagnosed. The most widely used metrics are described below~\cite{brownlee2016machine}.\\
\textbf{\textit{Accuracy (Acc)}:} The accuracy denotes total correctly identifying instances among all of the instances. Accuracy can be calculated using following formulas:
\begin{equation}
    ACC = \frac{T_{p}+ T_{N}}{T_{p}+T_{N}+F_{p}+F_{N}}
\end{equation}
\textbf{\textit{Precision ($P_n$:)}}
Precision is measured as the proportion of precisely predicted to all expected positive observations.
\begin{equation}
    P_n= \frac{T_p}{T_p+F_p}
\end{equation}
\textbf{\textit{Recall ($R_c$):}}
The proportion of overall relevant results that the algorithm properly recognizes is referred to as recall.
\begin{equation}
    \frac{T_p}{T_n+F_p}
\end{equation}
\textbf{\textit{Sensitivity ($S_n$)}:} Sensitivity denotes only true positive measure considering total instances and can be measured as follows:
\begin{equation}
    S_n= \frac{T_p}{T_p+F_N}
\end{equation}
\textbf{\textit{Specificity ($S_p$):}}  It identifies how many true negatives are appropriately identified and calculated as follows:
\begin{equation}
    S_p= \frac{T_N}{T_N+F_P}
\end{equation}
\textbf{\textit{F- measure:}} The F1 score is the mean of accuracy and recall in a harmonic manner. The highest f score is 1, indicating perfect precision and recall score.
\begin{equation}
 F-Measure=2\times\frac{\textrm{Precision}\times\textrm{ Recall}}{\textrm{Precision+Recall}}
\end{equation}
\textit{\textbf{Area under curve (AUC):}}
The area under the curve represents the models’ behaviors in different situations. The AUC can be calculated as follows:
\begin{equation}
    AUC= \frac{\sum R_i(I_p)- I_p((I_p +1)/2}{I_p+I_n}
\end{equation}
Where $l_p$  and $l_n$ denotes positive and negative data samples and $R_i$ is the rating of the $i^{th}$ positive samples.
\section{Article Selection}\label{article}
\subsection{Identification}
The Scopus and Web of Science (WOS) databases are utilized to find original research publications. Due to their high quality and peer review paper index, Scopus and WOS are prominent databases for article searching, as many academics and scholars utilized them for systematic review~\cite{malviya2015green,fahimnia2015green}. Keywords along with Boolean operator the title search was carried out as follows:\\
``disease" AND (``diagnsois" OR ``Supprot vector machine" OR ``SVM" OR ``KNN" OR ``K-nearest neighbor" OR ``logistic regression" OR ``K-means clustering" OR ``random forest" OR ``RF" OR ``adaboost" OR ``XGBoost" , ``decision tree" OR ``neural network" OR ``NN" OR ``artificial neural network" OR ``ANN" OR ``convolutional neural network" OR ``CNN" OR ``deep neural network" OR ``DNN" OR ``machine learning" or ``adversarial network" or ``GAN"). \\
The initial search yielded 16,209 and 2129 items, respectively, from Scopus and Web of Science (WOS).
\subsection{Screening}
Once the search period was narrowed to 2012-2021 and only peer-reviewed English papers were evaluated, the total number of articles decreased to 9117 for Scopus and 1803 for WOS, respectively. 
\subsection{Eligibility and Inclusion}
\begin{figure}[htbp]
    \centering
    \includegraphics[width=\textwidth]{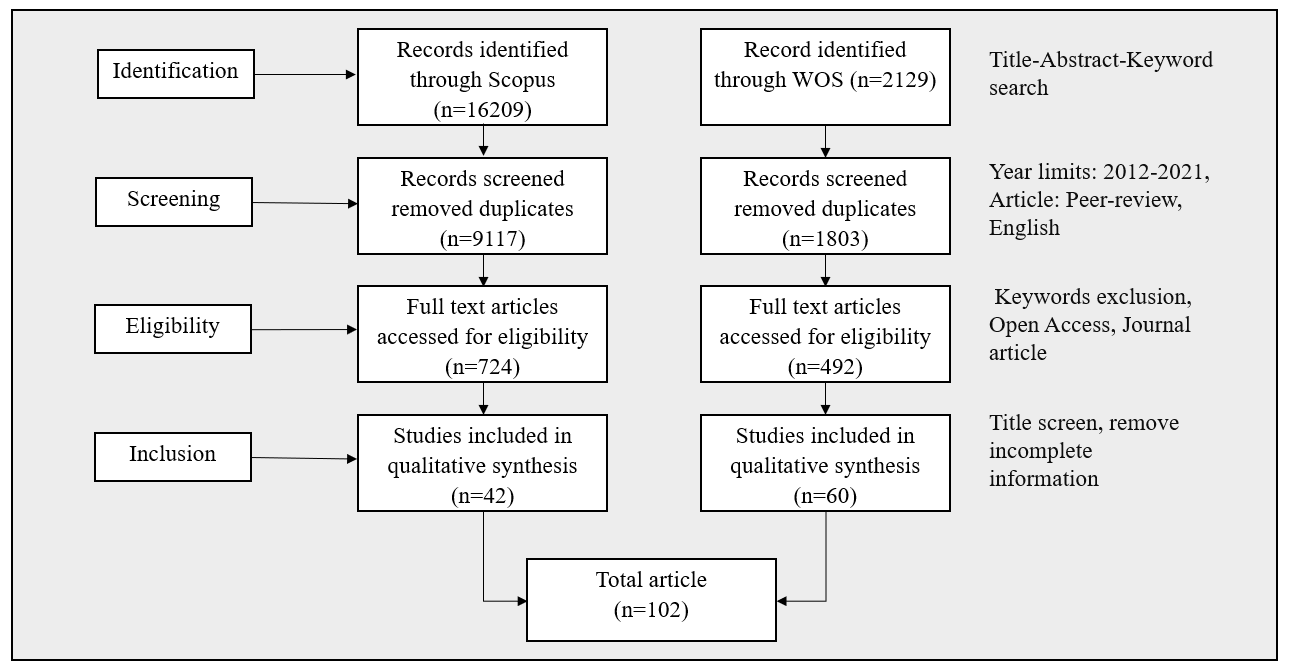}
    \caption{MLBDD article selection procedure used in this study.}
    \label{fig:fig4}
\end{figure}

These publications were chosen for further examination if they are open access and are journal articles. There were 1216 full-text articles (724 from the Scopus database and 492 from WOS). Bibliographic analysis was performed on all 1216 publications.
\\
One investigator (Z.S.) imported the 1216 article information as excel CSV data for future analysis. Excel duplication functions were used to identify and eliminate duplicates. Two independent reviewers (M.A. and Z.S.) examined the titles and abstracts of 1192 publications. Disagreements were settled through conversation. We omitted studies that were not relevant to Machine Learning but were relevant to disease diagnosis or vice versa.\\
After screening the titles and abstracts, the complete text of 102 papers was examined, and all 102 articles satisfied all inclusion requirements. Factors that contributed to the article's exclusion from the full-text screening includes:
\begin{enumerate}
    \item Inaccessibility of the entire text
    \item Non-human studies, book chapters, reviews
    \item Incomplete information related to test result
\end{enumerate}
Fig.~\ref{fig:fig4} shows the flow diagram of the systematic article selection procedure used in this study. 
\\

\section{Bibliometric Analysis}\label{bibliometric}
The bibliometric study in this section was carried out using reference literature gathered from the Scopus and WOS databases. 
The bibliometric study examines publications in terms of the subject area, co-occurrence network, year of publication, journal, citations, countries, and authors.
\subsection{Subject Area}
\begin{figure}[htbp]
    \centering
    \includegraphics[width=\textwidth]{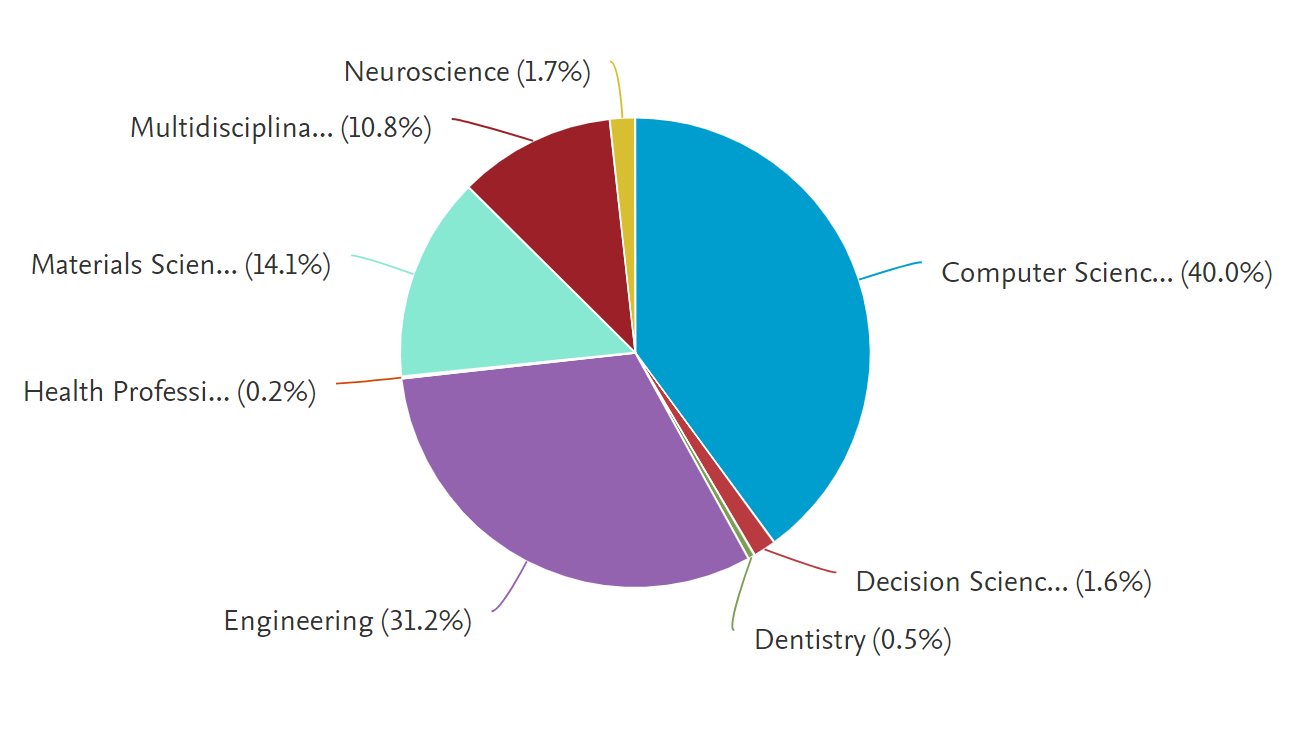}
    \caption{Distribution of articles by subject area.}
    \label{fig:fig5}
\end{figure}

Many research disciplines have uncovered machine learning-based disease diagnostics throughout the years. Figure~\ref{fig:fig5} depicts a schematic representation of machine learning-based disease detection spread across several research fields. According to the graph, computer science (40\%) and engineering (31.2\%) are two dominating fields that vigorously concentrated on MLBDD.
\subsection{Co-Occurrence Network}
\begin{figure}[htbp]
    \centering
    \includegraphics[width=\textwidth]{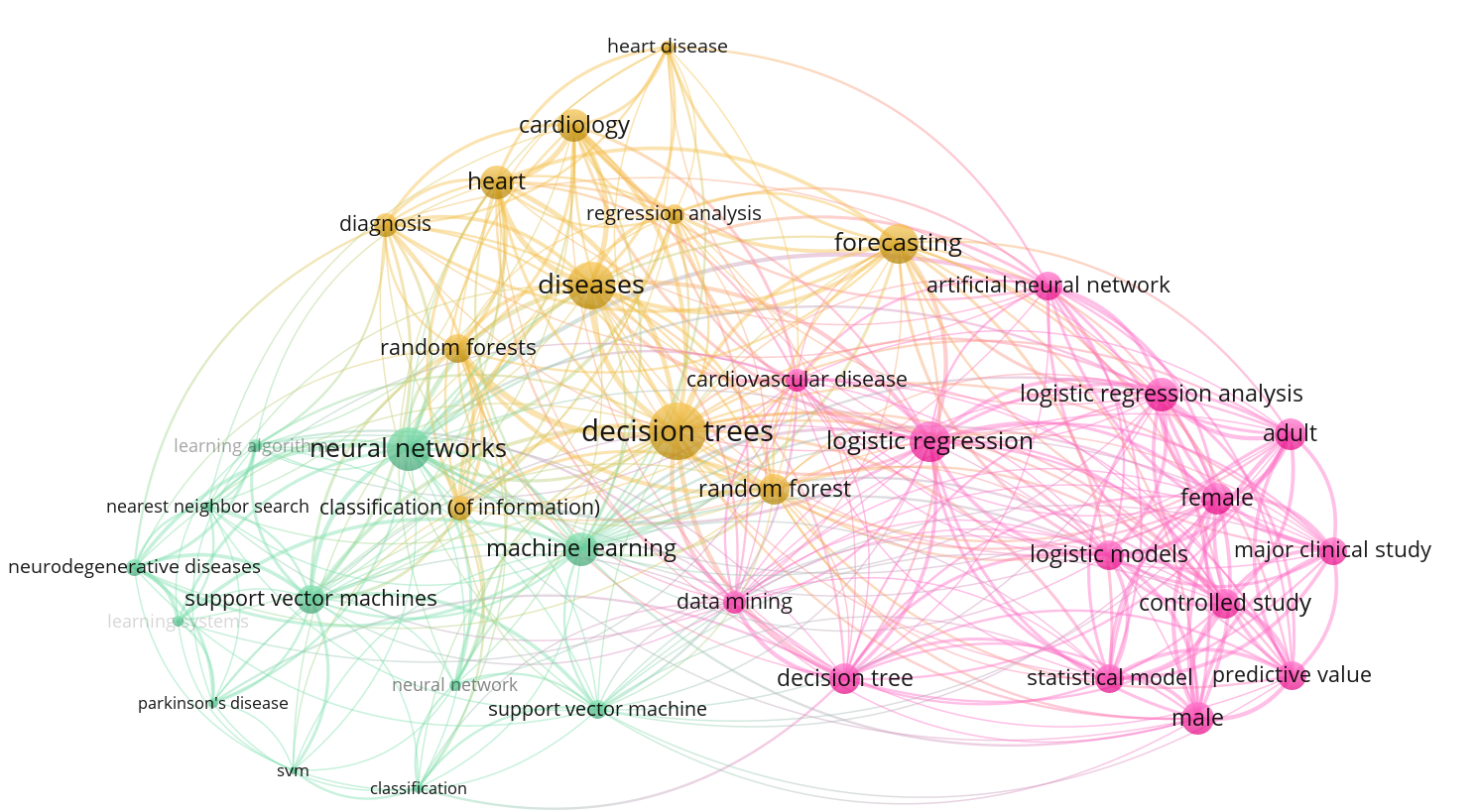}
    \caption{Bibliometric map representing co-occurrence analysis of keywords in network visualization.}
    \label{fig:figbib}
\end{figure}

Co-occurrence of keywords provides an overview of how the keywords are interconnected or used by the researchers. Fig.~\ref{fig:figbib} displays the co-occurrence network of the article's keywords and their connection, developed by VOSviewer software. The Figure shows that some of the significant clusters include Neural Networks (NN), Decision Trees (DT), Machine Learning (ML), and Logistic Regression (LR). Each cluster is also connected with other keywords that fall under that category. For instance, the NN cluster contains  Support Vector Machine (SVM), Parkinson's disease, and classification.
\subsection{Publication by Year}
\begin{figure}[htbp]
    \centering
    \includegraphics{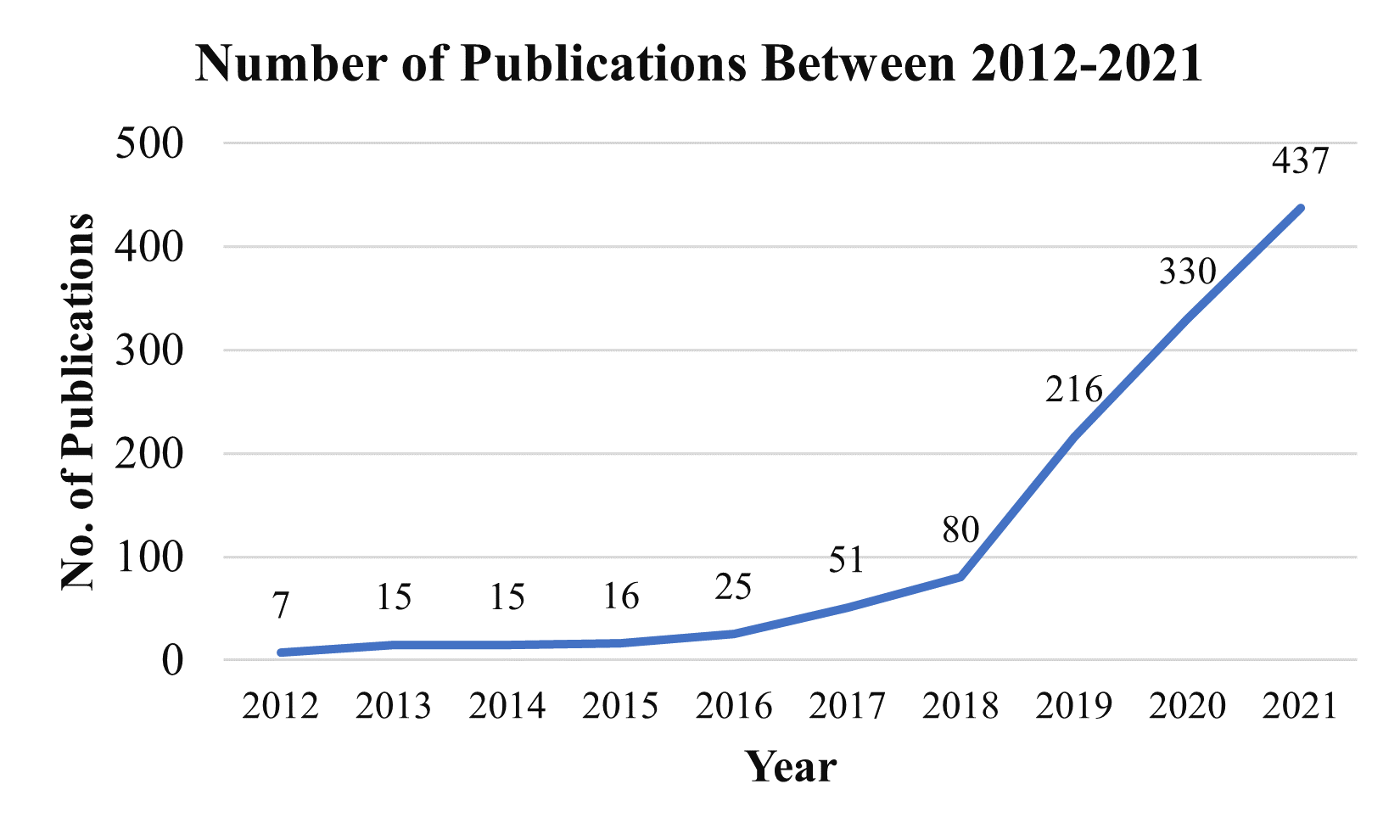}
    \caption{Publications of Machine Learning-based Disease Diagnosis (MLBDD) by year.}
    \label{fig:fig6}
\end{figure}

The exponential growth of journal publications is observed from 2017. Fig.~\ref{fig:fig6} displays the number of publications between 2012 to 2021 based on the Scopus and WOS data. Note that while the image may not accurately depict the MLBDD's real contribution, it does illustrate the influence of MLBDD over time. 
\subsection{Publication by Journal}
\begin{figure}[htbp]
    \centering
    \includegraphics[width=\textwidth]{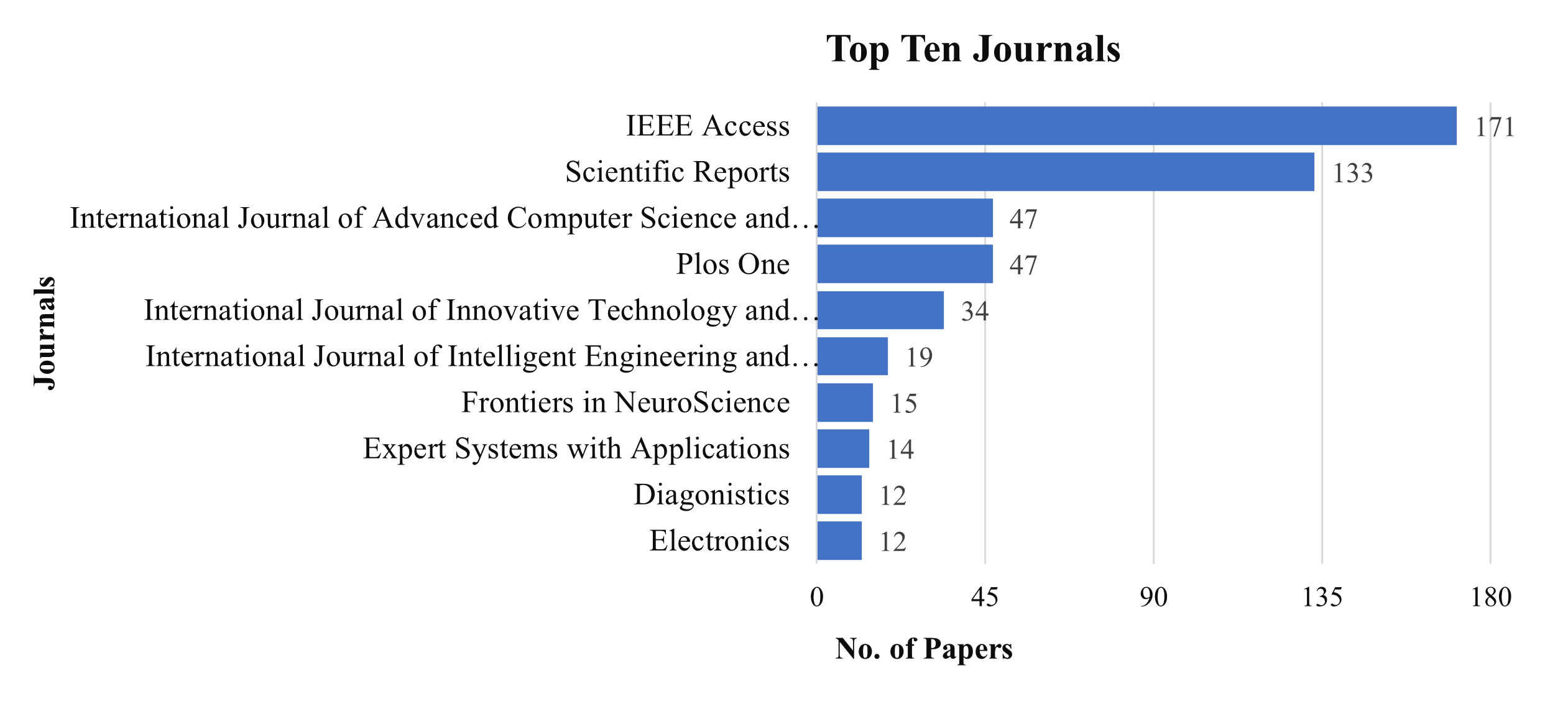}
    \caption{Publications by Journals}
    \label{fig:fig7}
\end{figure}

 We investigated the most prolific Journals in MLBDD domains based on our referred literature.The top ten journals and the number of articles published in the last ten years are depicted in Fig.~\ref{fig:fig7}. IEEE Access and Scientific Reports are the most productive journals that published 171 and 133 MLBDD articles respectively.
\subsection{Publication by Citations}
Citations are one of the primary indicators of an article's effect. Here we have identified the top ten cited articles using the R-studio tool. Table~\ref{tab:tab1} summarizes the top articles that achieved the highest citation during the year between 2012 to 2021. Note that, Google Scholar and other online databases may have various indexing procedures and times, therefore the citations in this manuscript may differ from the number of citations shown in this study. The Table shows that published articles by Motwani et al. (2017)~\cite{motwani2017machine} earned the most citations (257), with 51.4 citations per year, followed by Gray et al. (2013)~\cite{gray2013random}'s article, which obtained 218 citations. It is assumed that all the authors included in Table~\ref{tab:tab1} are among those prominent authors that contributed to MLBDD.
\begin{table}[htbp]
\caption{Top ten cited papers published in MLBDD in between 2012-2021 based on Scopus and WOS database.}

    \centering
    
    \begin{tabular}{p{.15\linewidth}p{.5\linewidth}p{.1\linewidth}}\toprule
         Author(s)&	\centering{Article Titles}&	Citation\\\midrule
         Motwani et al. (2017~)\cite{motwani2017machine}&	Machine Learning for prediction of all-cause mortality in patients with suspected coronary artery disease: a 5-year multicentre prospective registry analysis&
	257\\
Gray et al. (2013)~\cite{gray2013random}	&Random forest-based similarity measures for multi-modal classification of Alzheimer's disease&
	248\\
 Mohan et al. (2019)~\cite{mohan2019effective}&	Effective Heart disease prediction Using hybrid Machine Learning techniques&
	214\\
Yadav \& Jadvav (2019)~\cite{yadav2019deep}&	Deep Convolutional Neural Network based medical image classification for disease diagnosis
	&155\\
Zhang et al. (2015~)\cite{zhang2015detection}&	Detection of subjects and brain regions related to Alzheimer's disease using 3D MRI scans based on Eigenbrain and Machine Learning&
	147\\
Austin et al. (2013)~\cite{austin2013using}&Using methods from the data-mining and Machine-Learning literature for disease classification and prediction: a case study examining classification of Heart failure subtypes&139\\
	Sharmila \& Geethanjali (2016~)\cite{sharmila2016dwt}& DWT based detection of Epileptic Seizure From EEG signals using Naive Bayes and k-NN classifiers&134\\
	Lebedev et al. (2014)~\cite{lebedev2014random}&Random Forest ensembles for detection and prediction of Alzheimer's disease with a good between-cohort robustness&129\\
	Luz et al. (2013)~\cite{luz2013ecg}&ECG Arrhythmia classification based on optimum-path forest&111\\
	Challis et al. (2015)~\cite{challis2015gaussian}&Gaussian process classification of Alzheimer's disease and mild cognitive impairment from resting-state fMRI&107\\\bottomrule
    \end{tabular}
    
    \label{tab:tab1}
\end{table}
\subsection{Publication by Countries}
\begin{figure}[htbp]
    \centering
    \includegraphics[width=\textwidth]{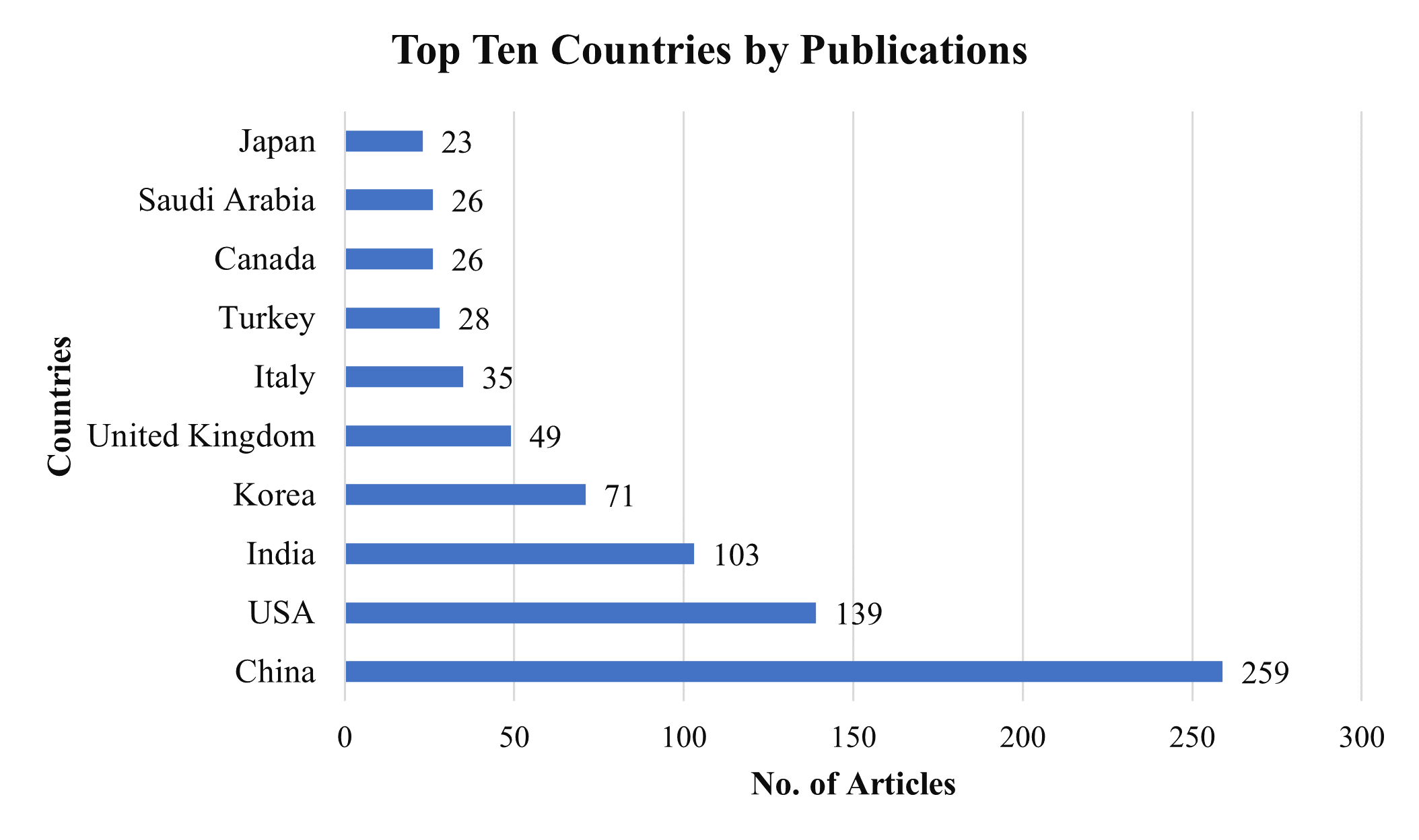}
    \caption{Top ten countries that contributed to MLBDD literature.}
    \label{fig:ttc}
\end{figure}

Fig.~\ref{fig:ttc} displayed that, China published the most publications in MLBDD, total 259 articles. USA and India are placed 2$^{nd}$ and 3$^{rd}$, respectively as they published 139 and 103 papers related to MLBDD. Interestingly, four out of the top ten productive countries are from Asia: China, India, Korea, and Japan.
\subsection{Publication by Author}
According to Table~\ref{tab:tta} Author Kim J published the most publications (20 out of 1216). Wang Y and Li J Ranked 2$^{nd}$ and 3$^{rd}$ by publishing 19 and 18 articles respectively. As shown in Table~\ref{tab:tta}, the number of papers produced by the top 10 authors ranges between 15-20. 
\begin{table}[htbp]
 \caption{Top ten authors based on total number of publications}
    \centering
    \begin{tabular}{cc}\toprule
         Author & Total Article\\\midrule
          Kim J & 20\\
         Wang Y & 19\\
         Li J & 18\\
         Liu Y & 18\\
         Chen Y & 17\\
         Kim H & 16\\
         Kim Y& 15\\
         Lee S & 15\\
         Li Y &15\\
         Wang L & 15\\\bottomrule
    \end{tabular}
    
    \label{tab:tta}
\end{table}
\section{Machine Learning Techniques for Different Disease Diagnosis}\label{ML}
Many academics and practitioners have used Machine Learning (ML) approaches in disease diagnosis. This section describes many types of Machine Learning-based Disease Diagnosis (MLBDD) that have received much attention because of their importance and severity. For example, due to the global relevance of COVID-19, several studies concentrated on COVID-19 disease detection using ML from 2020 to the present, which also received greater priority in our study. Severe diseases such as Heart disease, Kidney disease, Breast cancer, Diabetes, Parkinson's, Alzheimer's, and COVID-19 are discussed briefly, while other diseases are covered briefly under the "other disease."
\subsection{Heart Disease}
Most researchers and practitioners use Machine Learning (ML) approaches to identify cardiac disease~\cite{ansari2011automated,ahsan2021effect}. Ansari et al. (2011), for example, offered an automated coronary Heart disease diagnosis system based on Neuro-Fuzzy integrated systems that yield around 89\% accuracy~\cite{ansari2011automated}. One of the study's significant weaknesses is the lack of a clear explanation for how their proposed technique would work in various scenarios such as multiclass classification, big data analysis, and unbalanced class distribution. Furthermore, there is no explanation about the credibility of the model's accuracy, which has lately been highly encouraged in medical domains, particularly to assist users who are not from the medical domains in understanding the approach.\\
Rubin et al. (2017) uses Deep Convolutional Neural Network-based approaches to detect irregular cardiac sounds. 
The authors of this study adjusted the loss function to improve the training dataset's sensitivity and specificity. 
Their suggested model was tested in the 2016 PhysioNet computing competition. They finished second in the competition, with a final prediction of 0.95 specificity and 0.73 sensitivity~\cite{rubin2017recognizing}.\\
Aside from that, Deep Learning (DL)-based algorithms have lately received attention in detecting cardiac disease. 
Miao \& Miao et al. (2018), for example, offered a DL-based technique to diagnosing cardiotocographic fetal health based on a multiclass morphologic pattern. The created model is used to differentiate and categorize the morphologic pattern of individuals suffering from pregnancy complications. Their preliminary computational findings include accuracy of 88.02\%, a precision of 85.01\%, and an F-score of 0.85~\cite{miao2018cardiotocographic}. During that study, they employed multiple dropout strategies to address overfitting problems, which finally increased training time, which they acknowledged as a tradeoff for higher accuracy.\\
Although ML applications have been widely employed in Heart disease diagnosis, no research has been conducted that addressed the issues associated with unbalanced data with multiclass classification. 
Furthermore, the model's explainability during final prediction is lacking in most cases.
 Table~\ref{tab:tab2} summarizes some of the cited publications that employed ML and DL approaches in the diagnosis of cardiac disease. However, further information about Machine Learning-based cardiac disease diagnosis can be found in~\cite{ahsan2021machine}.
 \begin{table}[htbp]
 \caption{Referenced literature that considered Machine Learning based Heart disease diagnosis.}
    \centering
    \begin{tabular}{p{.1\linewidth}p{.15\linewidth}p{.2\linewidth}p{.1\linewidth}p{.1\linewidth}p{.2\linewidth}}\toprule
       Authors&	Contributions&	Algorithm&	Dataset&	Data type&	Performance evaluation \\\midrule 
Bemando et al. (2021~)\cite{bemando2021machine}&	Predict coronary heart disease&	Gaussian NB, Bernoulli NB and RF&	Cleveland dataset&	Tabular	&Accuracy- 85.00\%, 85.00\% and 75.00\%\\
Kumar \& Polepaka (2020)~\cite{kumar2020performance}&	Predicting heart diseases&	RF, CNN &Cleveland dataset&	Tabular&	RF (Accuracy- 80.327\%, Precision- 82\%, Recall- 80\%, F1-score- 80\%),
CNN (Accuracy- 78.688, Precision- 80\%, Recall- 79\%, F1-score- 78\%)\\
Sing et al. (2019)~\cite{singh2019multisurface}&	Heart disease classification&	SVM&	Cleveland database&	Tabular&	Accuracy- 73\%-91\%\\
Desai et al. (2019)~\cite{desai2019back}&Heart disease classification&	Back-propagation NN, LR&	Cleveland dataset&	Tabular&Accuracy (BNN- 85.074\%, LR- 92.58\%) \\
 Patil et al. (2018)~\cite{patil2018analysis}&	ECG arrhythmia for heart disease detection&	SVM and Cuckoo search optimized NN&
	Cleveland dataset&	Tabular& Accuracy (SVM-94.44\%)\\
Liu et al. (2012)~\cite{liu2012intelligent}&	Intelligent scoring system for the prediction of cardiac arrest within 72h&	SVM& Privately ownend&	Tabular&	Specificity- 78.8\%,
Sensitivity- 62.3\%,
Positive predictive value- 10\%, Negative predictive value- 98.2\%\\
Acharya et al. (2017)~\cite{acharya2017deep}&	Automatically identify 5 different categories of heartbeats in ECG signals&	CNN&

	MIT-BIH& Tabular&	Accuracy- 94\% (balance data)
Accuracy- 89.07\% (imbalance data)\\
Yang et al. (2018)~\cite{yang2018automatic}&	Novel heartbeat recognition method is presented	& SVM&MIT-BIH&
	Tabular&	Accuracy- 97.77\% (imbalance data),
Accuracy- 97.08\%( noise-free ECGs)\\\bottomrule
    \end{tabular}
    
    \label{tab:tab2}
\end{table}

\subsection{Kidney Disease}
Kidney disease, often known as Renal disease, refers to nephropathy or kidney damage. Patients with Kidney disease have decreased kidney functional activity, which can lead to kidney failure if not treated promptly. 
According to the National Kidney Foundation, 10\% of the world's population has chronic kidney disease (CKD), and millions die each year due to insufficient treatment.. The recent advancement of ML and DL-based kidney disease diagnosis may provide a possibility for those countries that are unable to handle the kidney disease diagnostic related tests~\cite{levey2012chronic}. For instance, 
Charleonnan et al. (2016) used publicly available datasets to evaluate four different ML algorithms: K-Nearest Neighbors (KNN), Support Vector Machine (SVM), Logistic Regression (LR), and Decision Tree classifiers and received the accuracy of 98.1\%, 98.3\%, 96.55\%, and 94.8\%, respectively~\cite{charleonnan2016predictive}. Aljaaf et al. (2018) conducted a similar study. 
The authors tested different ML algorithms, including RPART, SVM, LOGR, and MLP, using a comparable dataset, CKD, as used by~\cite{charleonnan2016predictive}, and found that MLP performed best (98.1 percent) in identifying chronic kidney disease~\cite{aljaaf2018early}. 
To identify chronic kidney disease, Ma et al. (2020)  utilizes a collection of datasets containing data from many sources~\cite{ma2020detection}. 
Their suggested Heterogeneous Modified Artificial Neural Network (HMANN) model obtained an accuracy of 87\%-99\%.

Table~\ref{tab:kidney} summarizes some of the cited publications that employed ML and DL approaches to diagnose kidney disease.
\begin{table}[htbp]
\caption{Referenced literature that considered machine learning based kidney disease diagnosis.}
    \centering
    \begin{tabular}{p{.1\linewidth}p{.2\linewidth}p{.2\linewidth}p{.15\linewidth}p{.08\linewidth}p{.1\linewidth}}\toprule
       Authors&	Contributions&	Algorithm&	Dataset&	Data type&	Performance evaluation \\\midrule
    Wasle et al. (2020)~\cite{walse2020effective}& Analysis of Chronic Kidney Disease&	NB, DT, and RF&	Chronic kidney disease dataset&	Tabular&	Accuracy- 100\% (RF)\\ 
       Nithya et al. (2020)~\cite{nithya2020kidney}& Kidney disease detection and segmentation& ANN and multi-kernel k-means clustering&	100 collected image data of patients	Ultrasound& Image&	Accuracy- 99.61\%\\
      Al Imran et al. (2018)~\cite{al2018classification}&Classification of Chronic kidney disease&	LR, Feedforward NN and Wide DL&	Chronic kidney disease dataset	&Tabular&	Feedforward NN (F1-score- 99\%, Precision- 97\%, Recall- 99\%, and AUC-99\%)\\
     Navaneeth \& Suchetha (2020~\cite{navaneeth2020dynamic}&Chronic kidney disease&	CNN-SVM	&Privately own dataset&	Tabular&	Accuracy-97.67\%,
Sensitivity– 97.5\%, Specificity- 97.83\%\\
Brunetti et al. (2019)~\cite{brunetti2019detection}&Detection and localization of kidneys in patients with autosomal dominant polycystic&CNN&	Privately own data&	Image&	Accuracy- 95\%\\\bottomrule

    \end{tabular}
    
    \label{tab:kidney}
\end{table}

\subsection{Breast Cancer}
Many scholars in the medical field have proposed Machine Learning (ML)-based Breast cancer analysis as a potential solution to early-stage diagnosis. Miranda \& Felipe (2015), for example, proposed Fuzzy Logic-based computer-aided diagnosis systems for breast cancer categorization. The advantage of Fuzzy Logic over other classic ML techniques is that it can minimize computational complexity while simulating the expert radiologist's reasoning and style. If the user inputs parameters such as contour, form, and density, the algorithm offers a cancer categorization based on their preferred method~\cite{miranda2015computer}. 
Miranda \& Felipe (2015)’s proposed model had an accuracy of roughly 83.34\%. The authors employed an approximately equal ratio of images for the experiment, which resulted in improved accuracy and unbiased performance. However, as the study did not examine the interpretation of their results in an explainable manner, it may be difficult to conclude that accuracy, in general, indicates true accuracy for both benign and malignant classifications. Furthermore, no confusion matrix is presented to demonstrate the models' actual prediction for the each class.\\
Zheng et al. (2014) presented hybrid strategies for diagnosing Breast cancer disease utilizing k-Means Clustering (KMC) and SVM. Their proposed model considerably decreased the dimensional difficulties and attained an accuracy of 97.38\% using Wisconsin Diagnostic Breast Cancer (WDBC) dataset~\cite{zheng2014breast}. The dataset is normally distributed and has 32 features divided into 10 categories. It is difficult to conclude that their suggested model will outperform in a dataset with an unequal class ratio, which may contain missing value as well.\\
To determine the best ML models, Asri et al. (2016) applied various ML approaches such as SVM, DT (C4.5), NB, and KNN on the Wisconsin Breast Cancer (WBC) datasets. According to their findings, SVM outperformed all other ML algorithms, obtaining an accuracy of 97.13\%~\cite{asri2016using}. However, if a same experiment is repeated in a different database, the results may differ. Furthermore, experimental results accompanied by ground truth values may provide a more precise estimate in determining which ML model is the best or not.\\
Mohammed et al. (2020) conducted a nearly identical study. The authors employ three ML algorithms to find the best ML methods: DT (J48), NB, and Sequential Minimal Optimization (SMO), and the experiment was conducted on two popular datasets: WBC and Breast Cancer dataset. One of the interesting aspects of this research is that they focused on data imbalance issues and minimized the imbalance problem through the use of resampling data labeling procedures. Their findings showed that the SMO algorithms exceeded the other two classifiers, attaining more than 95\% accuracy on both datasets~\cite{mohammed2020analysis}. However, in order to reduce the imbalance ratio, they used resampling procedures numerous times, potentially lowering the possibility of data diversity. As a result, the performance of those three ML methods may suffer on a dataset that is not normally distributed or imbalanced.\\
Assegie (2021) used the grid search approach to identify the best K-Nearest Neighbor (KNN) settings. Their investigation showed that parameter adjustment had a considerable impact on the model's performance. They demonstrated that by fine-tuning the settings, it is feasible to get 94.35\% accuracy, whereas the default KNN achieved around 90\% accuracy~\cite{assegie2021optimized}.\\
To detect breast cancer, Bhattacherjee et al. (2020) employed a Back-propagation Neural Network (BNN). The experiment was carried out in the WBC dataset with 9 features, and they achieved 99.27\% accuracy~\cite{bhattacherjee2020classification}.
Alshayeji et al. (2021) used the WBCD and WDBI datasets to develop a shallow- ANN model for classifying Breast cancer tumors. The authors demonstrated that the suggested model could classify tumors up to 99.85\% properly without selecting characteristics or tweaking the algorithms~\cite{alshayeji2022computer}.\\
Sultana et al. (2021) detect Breast cancer using a different ANN architecture on the WBC dataset. They employed a variety of NN architectures, including the Multilayer Perceptron (MLP) Neural Network, the Jordan/Elman NN, the Modular Neural Network (MNN), the Generalized Feed-Forward Neural Network (GFFNN), the Self-Organizing Feature Map (SOFM), the SVM Neural Network, the Probabilistic Neural Network (PNN), and the Recurrent Neural Network (RNN). Their final computational result demonstrates that the PNN with 98.24\% accuracy outperforms the other NN models utilized in that study~\cite{sultana2021early}. However, this study lacks the interpretability as of many other investigations because it does not indicate which features are most important during the prediction phase.\\
Deep Learning (DL) was also used by Ghosh et al. (2021). The WBC dataset was used by the authors to train seven deep learning (DL) models: ANN, CNN, GRU, LSTM, MLP, PNN, and RNN. Long short-term memory (LSTM) and Gated recurrent unit (GRU) demonstrated the best performance among all DL models, achieving an accuracy of roughly 99\%~\cite{ghosh2021performance}.
Table~\ref{tab:breast} summarizes some of the referenced literature that used ML and DL techniques in Breast cancer diagnosis.

\begin{table}[htbp]
\caption{Referenced literature that considered Machine Learning based Breast cancer disease diagnosis.}
    \centering
    \begin{tabular}{p{.1\linewidth}p{.15\linewidth}p{.2\linewidth}p{.1\linewidth}p{.1\linewidth}p{.2\linewidth}}\toprule
       Authors&	Contributions&	Algorithm&	Dataset&	Data type&	Performance evaluation \\\midrule
       Rajendran et al. (2020)~\cite{rajendran2020predicting}&Breast cancer&NB, BN, RF and DT (C4.5)&BCSC&	Image&  ROC - 0.937 (BN)\\
     Abdel et al. (2015)~\cite{abdel2015analysis}&Classification of breast density and mass &SVM&	Mini-MIAS, INBreast&	Image&	Mini-MIAS: - Accuracy- 99\%, AUC- 0.9325\\
       Sharma \& Khanna (2015)~\cite{sharma2015computer}& Classify vector features as malignant or non-malignant&SVM&IRMA ,DDSM& Image&	IRMA: Sensitivity- 99\%, Specificity- 99\% , DDSM: Sensitivity- 97\%, Specificity- 96\%\\
    Moon et al. (2017)~\cite{moon2017adaptive}&Classification of breast cancers by tumor size&LR - ANN&	156 Privately owned cases&	Image&	Accuracy- 81.8\%, Sensitivity- 85.4\%, Specificity- 77.8\%, AUC- 0.855\\
    Lo et al. (2016)~\cite{lo2016feasibility}&CAD tumor&	Binary-LR&
	18 Privately owned cases&	Image&	Accuracy- 80.39\%\\
	Venkatesh et al. (2015)~\cite{venkatesh2015going}&Differentiating malignant and benign masses& NB, LR - AdaBoost&	246 Privately owned image&	Image&	Sensitivity- 90\%, Specificity- 97.5\%, AUC- 0.98\\\bottomrule

    \end{tabular}
    
    \label{tab:breast}
\end{table}
\subsection{Diabetes}
According to the International Diabetes Federation (IDF), there are currently over 382 million individuals worldwide who have diabetes, with that number anticipated to increase to 629 million by 2045~\cite{naz2020deep}. Numerous studies widely presented ML-based systems for Diabetes patient detection. For example,kandhasamy \& Balamurali (2015) compared ML classifiers (J48 DT, KNN, RF, and SVM) for classifying patients with diabetes mellitus. The experiment was conducted on the UCI Diabetes dataset, and the KNN (K=1) and RF classifiers obtained near-perfect accuracy~\cite{kandhasamy2015performance}. However, one disadvantage of this work is that it used a simplified Diabetes dataset with only eight binary-classified parameters. As a result, getting 100\% accuracy with a less difficult dataset is unsurprising. Furthermore, there is no discussion of how the algorithms influence the final prediction or how the result should be viewed from a nontechnical position in the experiment.\\
Yahyaoui et al. (2019) presented a Clinical Decision Support Systems (CDSS) to aid physicians or practitioners with Diabetes diagnosis. To reach this goal, the study utilized a variety of ML techniques, including SVM, RF, and Deep Convolutional Neural Network (CNN). RF outperformed all other algorithms in their computations, obtaining an accuracy of 83.67\%, while DL and SVM scored 76.81\% and 65.38\% accuracy, respectively~\cite{yahyaoui2019decision}.\\
Naz \& Ahuja (2020) employed a variety of ML techniques, including Artificial neural networks (ANN), NB, DT, and DL, to analyze open-source PIMA Diabetes datasets. Their study indicates that DL is the most accurate method for detecting the development of Diabetes, with an accuracy of approximately 98.07\%~\cite{naz2020deep}. The PIMA dataset is one of the most thoroughly investigated and primary datasets, making it easy to perform conventional and sophisticated ML-based algorithms. As a result, gaining greater accuracy with the PIMA Indian dataset is not surprising. Furthermore, the paper makes no mention of interpretability issues and how the model would perform with an unbalanced dataset or one with a significant number of missing variables. As is widely recognized in healthcare, several types of data can be created that are not always labeled, categorized, and preprocessed in the same way as the PIMA Indian dataset. As a result, it is critical to examine the algorithms' fairness, unbiasedness, dependability, and interpretability while developing a CDSS, especially when a considerable amount of information is missing in a multiclass classification dataset.\\
Ashiquzzaman et al. (2017) developed a Deep Learning strategy to address the issue of overfitting in Diabetes datasets. The experiment was carried out on the PIMA Indian dataset and yielded an accuracy of 88.41\%. The authors claimed that performance improved significantly when dropout techniques were utilized and the overfitting problems were reduced~\cite{ashiquzzaman2017reduction}. Overuse of the dropout approach, on the other hand, lengthens overall training duration. As a result, as they did not address these concerns in their study, assessing whether their proposed model is optimum in terms of computational time is difficult.\\
Alhassan et al. (2018) introduced the King Abdullah International Research Center for Diabetes (KAIMRCD) dataset, which includes data from 14k people and is the world's largest Diabetic dataset. During that experiment, the author presented a CDSS architecture based on LSTM and GRU-based Deep Neural Networks, which obtained up to 97\% accuracy~\cite{alhassan2018type}.
Table~\ref{tab:diabetes} highlights some of the relevant publications that employed ML and DL approaches in the diagnosis of Diabetic disease.
\begin{table}[htbp]
    \caption{Referenced literature that considered Machine Learning based Diabetic disease diagnosis.}
    \centering
    \begin{tabular}{p{.1\linewidth}p{.15\linewidth}p{.2\linewidth}p{.1\linewidth}p{.1\linewidth}p{.2\linewidth}}\toprule
       Authors&	Contributions&	Algorithm&	Dataset&	Data type&	Performance evaluation \\\midrule
      
Fitriyani et al. (2019)~\cite{fitriyani2019development}&Diabetes and hypertension&

	DPM&	Privately owned&	Tabular& Accuracy- 96.74\%\\
Fernandez et al. (2021)~\cite{fernandez2021machine}&Type 1 diabetes&	RF&	DIABIM-MUNE&	Tabular&	AUC- 0.80\\
	Ali et al. (2020)~\cite{ali2020diabetes}&Diabetes classification&	KNN&	Privately owned- 4900 samples&	Tabular&	Accuracy- 99.9\%\\
	Tsao et al. (2018)~\cite{tsao2018predicting}&
Predict diabetic retinopathy and identify interpretable biomedical features&	SVM, DT, ANN, and LR&	Privately owned&	Tabular&	SVM (Accuracy- 79.5\%, AUC- 0.839)\\
Qtea et al. (2021)~\cite{qtea2021using}&Diabetes classification&	PSO and MLPNN& Privately owned&	Tabular&	Accuracy- 98.73\% \\\bottomrule
    \end{tabular}
    
    \label{tab:diabetes}
\end{table}
\subsection{Parkinson’s Disease}
Parkinson's disease is one of the conditions that has received a lot of attention in the ML literature. It is a slow-progressing chronic neurological disorder. When dopamine-producing neurons in certain parts of the brain are harmed or die, people have difficulty speaking, writing, walking, and doing other core activities~\cite{grover2018predicting}. There are several ML-based approaches have been proposed. For instance, Sriram et al. (2013) used KNN, SVM, NB, and RF algorithms to develop intelligent Parkinson's disease diagnosis systems. Their computational result shows that, among all other algorithms, RF shows the best performance (90.26\% accuracy), and NB demonstrate the worst performance (69.23\% accuracy)~\cite{sriram2013intelligent}.\\ 
Esmaeilzadeh et al. (2018) proposed a Deep CNN-based model to diagnose Parkinson's disease and achieved almost 100\% accuracy on train and test set~\cite{esmaeilzadeh2018end}. However, there was no mention of any overfitting difficulties in the trial. Furthermore, the experimental results do not provide a good interpretation of the final classification and regression, which is now widely expected, particularly in CDSS. Grover et al. (2018) also used DL-based approaches on UCI's Parkinson's telemonitoring voice dataset. Their experiment using DNN has achieved around 81.67\% accuracy in diagnosing patients with Parkinson's disease symptoms~\cite{grover2018predicting}.\\
Warjurkar \& Ridhorkar (2021) conducted a thorough study on the performance of the ML-based approach in decision support systems that can detect both Brain tumors and diagnose Parkinson's patients. Based on their findings, it was obvious that, when compared to other algorithms, Boosted Logistic Regression surpassed all other models, attaining 97.15\% accuracy in identifying Parkinson's disease patients. In tumor segmentation, however, the Markov random technique performed best, obtaining an accuracy of 97.4\%~\cite{warjurkar2021study}.
Parkinson's disease diagnosis using ML and DL approaches is summarized in Table~\ref{tab:parkinson} , which includes a number of references to the relevant research.
\begin{table}[htbp]
 \caption{Referenced literature that considered Machine Learning based parkinson disease diagnosis.}
    \centering
    \begin{tabular}{p{.15\linewidth}p{.15\linewidth}p{.1\linewidth}p{.1\linewidth}p{.1\linewidth}p{.2\linewidth}}\toprule
       Authors&	Contributions&	Algorithm&	Dataset&	Data type&	Performance evaluation \\\midrule
     Sherly et al. (2021)~\cite{sherly2021novel}& Parkinson’s disease&	KMC and DT&	Privately owned&	Speech&
	Accuracy- 95.56\%\\
Nurrohman et al. (2021)~\cite{nurrohman2020parkinson}&Parkinson's disease subtype classification&	DT, LR &	PPMI&	Tabular&	Accuracy- 98.3\%, Sensitivity- 98.41\%, and Specificity- 99.07\%\\
	Asmae et al. (2020)\cite{asmae2020parkinson}&Parkinson's disease identification&	KNN and ANN&	Parkinson's  UI  machine learning  dataset&	Tabular&
	ANN (Accuracy- 96.7\%) \\
	Guruler et al. (2017)~\cite{guruler2017novel}&Diagnosis system for Parkinson’s disease& ANN, KMC&	Parkinsons dataset&	Speech and sound& Accuracy- 99.52\%\\
	Shetty et al. (2016)~\cite{shetty2016svm}&identify Parkinson's disease&	SVM& NIHS&	Speech and sound& Accuracy- 83.33\%, True positive- 75\%, False positive- 16.67\%\\\bottomrule
    \end{tabular}
   
    \label{tab:parkinson}
\end{table}
\subsection{COVID-19}
The new Severe Acute Respiratory Syndrome Coronavirus 2 (SARS-CoV-2), also known as COVID-19, pandemic has become humanity's greatest challenge in contemporary history. Despite the fact that a vaccine had been advanced in distribution because to the global emergency, it was unavailable to the majority of people for the duration of the crisis~\cite{ahsan2021detection}. Because of the new COVID-19 Omicron strain's high transmission rates and vaccine-related resistance, there is an extra layer of concern. The gold standard for diagnosing COVID-19 infection is now Real-Time Reverse Transcription-Polymerase Chain Reaction (RT-PCR)~\cite{haghanifar2020covid,tahamtan2020real}. Throughout the epidemic, the researcher advocated other technologies including as chest X-rays and Computed Tomography (CT) combined with Machine Learning and Artificial Intelligence to aid in the early detection of people who might be infected. For example, Chen et al. (2020) proposed a UNet++ model employing CT images from 51 COVID-19 and 82 non-COVID-19 patients and achieved an accuracy of 98.5\%~\cite{chen2020deep}.
Ardakani et al. (2020) used a small dataset of 108 COVID-19 and 86 non-COVID-19 patients to evaluate ten different DL models and achieved a 99\% overall accuracy~\cite{ardakani2020application}. Wang et al. (2020) built an inception-based model with a large dataset, containing 453 CT scan images, and achieved 73.1\% accuracy. However, the model's network activity and region of interest were poorly explained~\cite{wang2020covid} . Li et al. (2020) suggested the COVNet model and obtain 96\% accuracy utilizing a large dataset of 4356 chest CT images of Pneumonia patients, 1296 of which were verified COVID-19 cases~\cite{li2020artificial}.\\
Several studies investigated and advised screening COVID-19 patients utilizing chest X-ray images in parallel, with major contributions in~\cite{hemdan2020covidx,sethy2020detection,narin2021automatic}. For example,Hemdan et al. (2020) used a small dataset of only 50 images to identify COVID-19 patients from chest X-ray images with an accuracy of 90\% and 95\%, respectively, using VGG19 and ResNet50 models~\cite{hemdan2020covidx}. Using a dataset of 100 chest X-ray images, Narin et al. (2021) distinguished COVID-19 patients from those with Pneumonia with 86\% accuracy~\cite{narin2021automatic}.\\
In addition, in order to develop more robust and better screening systems, other studies considered larger datasets. For example, Brunese et al. (2020) employed 6505 images with a data ratio of 1:1.17, with 3003 images classified as COVID-19 symptoms and 3520 as "other patients" for the objectives of that study~\cite{brunese2020explainable}. With a dataset of 5941 images, Ghoshal \& Tucker (2020) achieved 92.9\% accuracy~\cite{ghoshal2020estimating}. However, neither study looked at how their proposed models would work with data that was severely unbalanced and had mismatched class ratios. Apostolopoulos \& Mpesiana (2020) employed a CNN-based Xception model on an imbalanced dataset of 284 COVID-19 and 967 non-COVID-19 patient chest X-ray images and achieved 89.6\% accuracy~\cite{apostolopoulos2020covid}.\\
 The following Table~\ref{tab:covid} summarizes some of the relevant literature that employed ML and DL approaches to diagnose COVID-19 disease.
 \begin{table}
     \caption{Referenced literature that considered Machine Learning based COVID-19 disease diagnosis.}
    \centering
    \begin{tabular}{p{.1\linewidth}p{.15\linewidth}p{.2\linewidth}p{.1\linewidth}p{.1\linewidth}p{.2\linewidth}}\toprule
       Authors&	Contributions&	Algorithm&	Dataset&	Data type&	Performance evaluation \\\midrule
      Li et al. (2020)~\cite{li2020artificial}& COVID-19 disease detection&	CNN&	Mixed dataset&	Image&	Accuracy- 90\%\\
      Chen et al. (2020)~\cite{chen2020deep}& COVID-19 disease detection&	CNN&	Mixed dataset&	Image& Accuracy- 98.5\%\\
      Song et al. (2021)~\cite{song2021deep}&COVID-19 disease detection&	CNN&	Mixed dataset&	Image&	Accuracy- 86\%\\
      Jin et al. (2021)~\cite{jin2020development}&COVID-19 disease detection& CNN& Cohen’s dataset&	Image& Accuracy- 94.1\%\\
       Ahsan et al. (2020)~\cite{ahsan2020deep}& COVID-19 disease detection and image segmentation&	CNN&	Cohen’s dataset&	Image and Tabular&	Accuracy- 95.38\%\\\bottomrule
        
    \end{tabular}

    \label{tab:covid}
\end{table}
\subsection{Alzheimer’s Disease}
Alzheimer is a brain illness that often begins slowly but progresses over time, and it affects 60\%-70\% of those who are diagnosed with dementia~\cite{graham2009alzheimer}. Alzheimer's disease symptoms include language problems, confusion, mood changes, and other behavioral disorders. Body functions gradually deteriorated, and the usual life expectancy is three to nine years after diagnosis. Early diagnosis, on the other hand, may assist to avoid and take required actions to enter into suitable treatment as soon as possible, which will also raise the possibility of life expectancy. Machine Learning and Deep Learning have shown promising outcomes in detecting Alzheimer's disease patients throughout the years. For instance,Neelaveni \& Devasana (2020) proposed a model that can detect Alzheimer patients using SVM and DT, and achieved an accuracy of  85\% and 83\% respectively~\cite{neelaveni2020alzheimer}. Collij et al. (2016) also used SVM to detect single-subject Alzheimer’s disease and Mild Cognitive Impairment (MCI) prediction and achieved an accuracy of 82\%~\cite{collij2016application}.\\
Multiple algorithms have been adopted and tested in developing ML based Alzheimer disease diagnosis. For example, Vidushi \& Shrivastava (2019) experimented using Logistic Regression (LR), SVM, DT, ensemble Random Forest (RF), and Boosting Adaboost and achieved an accuracy of 78.95\%, 81.58\%, 81.58\%, 84.21\%, and 84.21\% respectively~\cite{vidushi2019diagnosis}.
Many of the study adopted CNN based approach to detect Alzheimer patients as CNN demonstrates robust results in image processing compared to other existing algorithms. As a consequence, Ahmed et al. (2020) proposed a CNN model for earlier diagnosis and classification of Alzheimer disease. Within the dataset consists of 6628 MRI images, the proposed model achieved 99\% accuracy~\cite{ahmed2020ensemble}. Nawaz et al. (2020) proposed deep feature-based models and achieved an accuracy of 99.12\%~\cite{nawaz2020deep}. Additionally, Studies conducted by Haft-Javaherian et al. (2019)~\cite{haft2019deep} and Aderghal et al. (2017)~\cite{aderghal2017classification} are some of the CNN based study that also demonstrates the robustness of CNN based approach in Alzheimer disease diagnosis.
ML and DL approaches employed in the diagnosis of Alzheimer's disease are summarized in Table~\ref{tab:alzheimer}.
\begin{table}
    \caption{Referenced literature that considered Machine Learning based Alzheimer disease diagnosis.}
    \centering
    \begin{tabular}{p{.1\linewidth}p{.15\linewidth}p{.2\linewidth}p{.1\linewidth}p{.1\linewidth}p{.2\linewidth}}\toprule
       Authors&	Contributions&	Algorithm&	Dataset&	Data type&	Performance evaluation \\\midrule
      Sun et al. (2021)~\cite{sun2021automatic}&Automatic diagnosis of alzheimer's disease and mild cognitive impairment&	CNN+SVM& F-FDG PET :PET&	Image&	Accuracy- 74\%-90\%\\
      Kunag et al. (2021)~\cite{kuang2021prediction}&
Predicting transition from mild cognitive impairment to alzheimer's& LR,ARN,DT&
	1913 privately owned cases&	Tabular&	Accuracy- (89.52 ± 0.36\%), AUC-ROC (92.08 ± 0.12), Sensitivity- (82.11 ± 0.42\%) and Postivie predictive value (75.26 ± 0.86\%)\\
	Manzak et al. (2019)~\cite{manzak2019automated}& Automatic classification of alzheimer's&	DNN+ RF&	Tabular&	Accuracy- 67\%\\\bottomrule

    \end{tabular}
    
    \label{tab:alzheimer}
\end{table}
\subsection{Other Diseases}
Beyond the disease mentioned above, ML and DL have been used to identify various other diseases. Big data and increasing computer processing power are two key reasons for this increased use. For example, Mao et al. (2020) used Decision Tree (DT) and Random Forest (RF) to disease classification based on eye movement~\cite{mao2020disease}. Nosseir \& Shawky (2019) evaluated KNN and SVM to develop automatic skin disease classification systems, and the best performance was observed using KNN by achieving an accuracy of 98.22\%~\cite{nosseir2019automatic}. Khan et al. (2020) employed CNN-based approaches such as VGG16 and VGG19 to classify multimodal Brain tumors. The experiment was carried out using publicly available three image datasets:  BraTs2015, BraTs2017, and BraTs2018, and achieved 97.8\%, 96.9\%, and 92.5\% accuracy, respectively~\cite{khan2020multimodal}. Amin et al. (2018) conducted a similar experiment utilizing the RF classifier for tumor segmentation. The authors achieved 98.7\%, 98.7\%, 98.4\%, 90.2\%, and 90.2\% accuracy using BRATS 2012, BRATS 2013, BRATS 2014, BRATS 2015, and ISLES 2015 dataset, respectively~\cite{amin2018detection}.\\
Dai et al. (2019) proposed a CNN-based model to develop an application to detect Skin cancer. The authors used a publicly available dataset, HAM10000, to experiment and achieved 75.2\% accuracy~\cite{dai2019machine} . Daghrir et al. (2020) evaluated KNN, SVM, CNN, Majority Voting using ISIC (International Skin Imaging Collaboration) dataset to detect Melanoma skin cancer. The best result was found using Majority Voting (88.4\% accuracy)~\cite{daghrir2020melanoma}.
Table~\ref{tab:other} summarizes some of the referenced literature that used ML and DL techniques in various disease diagnosis.

\begin{table}
     \caption{Referenced literature that considered Machine Learning on various disease diagnosis.}
    \centering
    \begin{tabular}{p{.1\linewidth}p{.15\linewidth}p{.2\linewidth}p{.1\linewidth}p{.1\linewidth}p{.2\linewidth}}\toprule
       Authors&	Contributions&	Algorithm&	Dataset&	Data type&	Performance evaluation \\\midrule
     Dhaliwal et al. (2021)~\cite{dhaliwal2021accurate}&Classify pediatric colonic inflammatory bowel disease subtype&	RF&	74 Privately owned cases&	Image&	Accuracy- 100\%\\
       Fathi et al. (2020)~\cite{fathi2020machine}&classification of liver diseases&	svm&	ILPD and BUPA&	Tabular	&Accuracy- 90\%-92\%,
Sensitivity- 89\%-91\%,
F1-score- 94\%-94.3\%\\
Wang et al. (2014)~\cite{wang2014logistic}&Hypertension& LR and ANN&	BRFSS&	Tabular&	Accuracy- 72\%,
AUC> 0.77\\
Kalaiselvi et al. (2020)~\cite{kalaiselvi2020deriving}&Brain tumor diagnostic&	CNN	& Brain tumor challenge websites and MRI centers&	Image&	Accuracy- 90\%-99\%\\
Usman \& Rajpoot (2017)~\cite{usman2017brain}& Brain tumor segmentation
for multi-modality MRI&	RF&	MICCAI, BraTS 2013&	Image&	88\% disc overlap\\
Waheed et al. (2017)~\cite{waheed2017efficient}&Melanoma detection with dermoscopic images&	SVM with color and feature extractor&	PH2& 	Image&	Accuracy- 96\%\\
    Kamboj et al. (2018)~\cite{kamboj2018color}&Melanoma skin cancer detection&
	NB, DT, and KNN&	MED-NODE&	Image& DT (Accuracy- 82.35\%) \\
Magalhaes et al. (2021)~\cite{magalhaes2021comparison}&Skin cancer detection with infrared thermal imaging&	Ensemble learning and DL& &	Image&	Precision- 0.9665, Recall- 0.9411, F1-score- 0.9536,  ROC-AUC- 0.9185\\
Chen et al. (2020)~\cite{chen2020classification}& Hepatocellular carcinoma&	InceptionV3&	Genomic data commons databases&	Image&	Accuracy- 89\%-96\%\\
Das et al. (2019)~\cite{das2019deep}& Identification of liver cancer&	Watershed gaussian based DL (WGDL)& Privately owned&	Image&	Accuracy- 99.38\%\\
Wang et al. (2021)~\cite{wang2021predicting}& Hepatocellular carcinoma (HCC) postoperative death outcomes&	RF,Gradient boosting,Gbm,
LR,DT&	BioStudies database&Tabular	&	AUC- 0.803 (RF)\\\bottomrule
    \end{tabular}
   
    \label{tab:other}
\end{table}
\section{Algorithm and Dataset Analysis}\label{AD}
\begin{figure}[htbp]
    \centering
    \includegraphics[width=\textwidth]{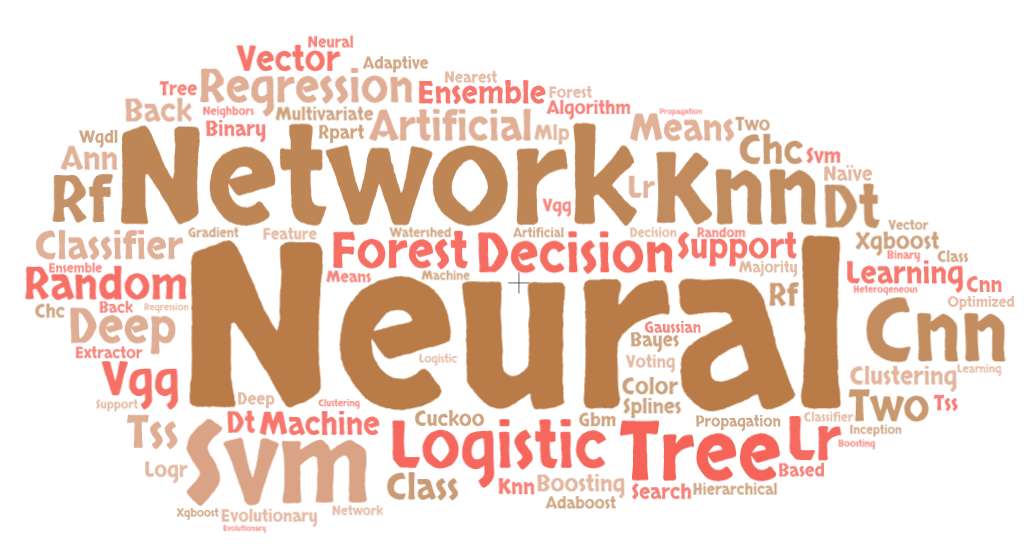}
    \caption{Word cloud for most frequently used ML algorithms in MLBDD publications.}
    \label{fig:wordcloud}
\end{figure}

Most of the referenced literature considered multiple algorithms in MLBDD approaches. Here we have addressed multiple algorithms as hybrid approaches. For instance, Sun et al. (2021) used hybrid approaches to predict coronary Heart disease using Gaussian Naïve Bayes, Bernoulli Naïve Bayes, and Random Forest (RF) algorithms~\cite{sun2021automatic}. Bemando et al. (2021) adopted CNN and SVM to automate the diagnosis of Alzheimer's disease and mild cognitive impairment~\cite{bemando2021machine}. Saxena et al. (2019) used KNN and Decision Tree (DT) in Heart disease diagnosis~\cite{saxena2019heart}; Elsalamony (2018) employed Neural Networks (NN) and SVM in detecting  Anaemia disease in human red blood cells~\cite{elsalamony2018detection}. One of the key benefits of using the hybrid technique is that it is more accurate than using single ML models.\\
 According to the relevant literature, the most extensively utilized induvial algorithms in developing MLBDD models are CNN, SVM, and LR. For instance, Kalaiselvi et al. (2020) proposed CNN based approach in Brain tumor diagnosis~\cite{kalaiselvi2020deriving}; Dai et al. (2019) used CNN in developing a device inference app for Skin cancer detection~\cite{dai2019machine};Fathi et al. (2020) used SVM to classify liver diseases~\cite{fathi2020machine}; Sing et al. (2019) used SVM to classify the patients with Heart disease symptoms~\cite{singh2019multisurface};and Basheer et al. (2019) used Logistic Regression to detect Heart disease~\cite{basheer2019ensembling}.\\
 Fig.~\ref{fig:wordcloud} depicts the most commonly used Machine Learning algorithms in disease diagnosis. The bolder and larger font emphasizes the importance and frequency with which the algorithms in MLBDD are used. 
Based on the Figure, we can observe that Neural Networks, CNN, SVM, and Logistic Regression are the most commonly employed algorithms by MLBDD researchers.
\\
Most MLBDD researchers utilize publically accessible datasets since they do not require permission and provide sufficient information to do the entire study. Manually gathering data from patients, on the other hand, is time-consuming; yet, numerous research utilized privately collected/owned data, either owing to their special necessity based on their experiment setup or to produce a result with actual data~\cite{liu2012intelligent,navaneeth2020dynamic,brunetti2019detection,moon2017adaptive,venkatesh2015going}. The Cleveland Heart disease dataset, PIMA dataset, and Parkinson dataset are the most often utilized datasets in disease diagnosis areas. Table~\ref{tab:datasource} lists publicly available datasets and sources that may be useful to future academics and practitioners.
\begin{table}[htbp]
    \centering
    \caption{Most widely used disease diagnosis dataset URL along with the referenced literature.}
    \vspace{.5 cm}
    \begin{tabular}{p{.2\linewidth}p{.1\linewidth}p{.1\linewidth}p{.3\linewidth}}\toprule
    Study&	Disease&	Dataset&	URL\\\midrule
        \cite{bemando2021machine,bharti2021prediction,saw2020estimation,kumar2020performance,singh2019multisurface,desai2019back,patil2018analysis}&Heart disease&	Cleveland database&	\url{https://archive.ics.uci.edu/ml/datasets/heart+disease}\\
        \cite{walse2020effective,al2018classification,charleonnan2016predictive,aljaaf2018early}& Kidney disease&	Chronic kidney disease dataset&	\url{https://archive.ics.uci.edu/ml/datasets/Chronic_Kidney_Disease}\\
        \cite{gill2016computational,kandhasamy2015performance,naz2020deep,ashiquzzaman2017reduction}&Diabetics&	Pima diabetic dataset&	\url{https://www.kaggle.com/uciml/pima-indians-diabetes-database}\\
        \cite{guruler2017novel,asmae2020parkinson,nurrohman2020parkinson,sriram2013intelligent}&Parkinson disease&	Parkinsons Dataset&	\url{https://archive.ics.uci.edu/ml/datasets/parkinsons}\\
        \cite{zheng2014breast,asri2016using,mohammed2020analysis}&Breast cancer&	WDBC dataset&	\url{https://archive.ics.uci.edu/ml/datasets/Breast+Cancer+Wisconsin+(Diagnostic)}\\
        \cite{ahsan2021detecting,narin2021automatic}&COVID-19&Covid-chest X-ray  dataset& \url{https://github.com/ieee8023/covid-chestxray-dataset}\\\bottomrule
        
    \end{tabular}
    
    \label{tab:datasource}
\end{table}
\section{Discussion}\label{discussion}
In the last 10 years, Machine Learning (ML) and Deep Learning (DL) have grown in prominence in disease diagnosis, which the annotated literature has strengthened in this study. 
The review began with specific research questions and attempted to answer them using the reference literature. 
According to the overall research, CNN is one of the most emerging algorithms, outperforming all other ML algorithms due to its solid performance with both image and tabular data~\cite{li2020artificial,chen2020classification, sun2012novel,kalaiselvi2020deriving}. 
Transfer learning is also gaining popularity since it does not necessitate constructing a CNN model from scratch and produces better results than typical ML methods~\cite{chen2020deep,acharya2017deep}. 
Aside from CNN, the reference literature lists SVM, RF, and DT as some of the most common algorithms utilized widely in MLBDD. 
Furthermore, several researchers are emphasizing ensemble techniques in MLBDD~\cite{magalhaes2021comparison,wang2021predicting}. 
Nonetheless, when compared to other ML algorithms, CNN is the most dominating. VGG16, VGG19, ResNet50, and UNet++ are among of the most prominent CNN architectures utilized widely in disease diagnosis.\\
In terms of databases, it was discovered that UCI repository data is the preferred option of academics and practitioners for constructing a Machine Learning-based Disease Diagnosis (MLBDD) model. 
However, while the current dataset frequently has shortcomings, several researchers have recently relied on additional data acquired from the hospital or clinic (i.e., imbalance data, missing data). 
To assist future researchers and practitioners interested in studying MLBDD, we have included a list of some of the most common datasets utilized in the reference literature in Table~\ref{tab:datasource}, along with the link to the repository.\\
As previously indicated, there were several inconsistencies in terms of assessment measures published by the literature. For example, some research reported their results with accuracy~\cite{patil2018analysis}; others provided with accuracy, precision, recall, and F1-score~\cite{kumar2020performance}; while a few studies emphasized sensitivity, specificity, and true positive~\cite{sharma2015computer}. As a result, there were no criteria for the authors to follow in order to report their findings correctly and genuinely. Nonetheless, of all assessment criteria, accuracy is the most extensively utilized and recognized by academics.\\
With the emergence of COVID-19, MLBDD research switched mostly on Pneumonia and COVID-19 patient detection beginning in 2020, and COVID-19 remains a popular subject as the globe continues to battle this disease. 
As a result, it is projected that the application of ML and DL in the medical sphere for disease diagnosis would expand significantly in this domain in the future as well. Many questions have been raised due to the progress of ML and DL-based disease diagnosis. 
For example, if a doctor or other health practitioner incorrectly diagnoses a patient, he or she will be held accountable. 
However, if the machine does, who will be held accountable? Furthermore, fairness is an issue in ML because most ML models are skewed towards the majority class. As a result, future research should concentrate on ML ethics and fairness.\\
Model interpretation is absent in nearly all investigations, which is surprising. Interpreting machine learning models used to be difficult, but explainable and interpretable XAI have made it much easier. 
Despite the fact that the previous MLBDD lacked sufficient interpretations, it is projected that future researchers and practitioners would devote more attention to interpreting the machine learning model due to the growing demand for model interpretability.\\
The idea that ML alone will enough to construct an MLBDD model is a flawed one. To make the MLBDD model more dynamic, it may be anticipated that the model will need to be developed and stored on a cloud system, as the heath care industry generates a lot of data that is typically kept in cloud systems. As a result, the adversarial attack will focus on patients' data, which is very sensitive. For future ML-based models, the data bridge and security challenges must be taken into consideration.\\
It is a major issue to analyze data if there is a large disparity in the data. As the ML-based diagnostic model deals with human life, every misdiagnosis is a possible danger to one's health. 
However, despite the fact that many study used the imbalance dataset to perform their experiment, none of the cited literature highlights issues related to imbalance data. Thus, future work should demonstrate the validity of any ML models while developing with imbalanced data.\\
Within the many scopes this review paper also have some limitations which can be summarized as follows:
\begin{enumerate}
    \item The study first searched the Scopus and WOS databases for relevant papers and then examined other papers that were pertinent to this investigation. If other databases like Google Scholar and Pubmed were used, the findings might be somewhat different. As a result, our study may provide some insight into MLBDD, but there is still a great deal of information that is outside of our control.
    \item ML algorithms, DL algorithms, dataset, disease classifications, and evaluation metrics are highlighted in the review. Though the suggested ML process is thoroughly examined in reference literature, this paper does not go into that level of detail.
    \item  Only those publications that adhered to a systematic literature review technique were included in the study's paper selection process. 
Using a more comprehensive range of keywords, on the other hand, might lead to higher search activity. However, our SLR approach will provide researchers and practitioners with a more thorough understanding of MLBDD.
\end{enumerate}
\section{Conclusions}\label{conclusion}
This study reviewed the papers published between 2012-2021 that focused on Machine Learning-based Disease Diagnosis (MLBDD). Researchers are particularly interested in some diseases, such as Heart disease, Breast cancer, Kidney disease, Diabetes, Alzheimer's, and Parkinson's diseases, which are discussed considering machine learning/deep learning-based techniques. Additionally, some other ML-based disease diagnosis approaches are discussed as well. Prior to that, 
A bibliometric analysis was performed, taking into account a variety of parameters such as subject area, publication year, journal, country, and identifying the most prominent contributors in the MLBDD field. 
Because of its remarkable performance in constructing a robust model, our major conclusion implies that the Convolution Neural Network (CNN) is the most popular method for researchers. We anticipate that our review will guide both novice and expert research and practitioners in MLBDD. It would be interesting to see some research work based on the limitations addressed in the discussion section. Additionally, future works in MLBDD might focus on multiclass classification with highly imbalanced data along with highly missing data, explanation and Interpretation of multiclass data classification using XAI, and optimizing the big data containing numerical, categorical, and image data.  
\bibliographystyle{unsrt}  
\bibliography{main}

\end{document}